\documentclass{article}

 \usepackage[preprint]{neurips_2025}

\usepackage[utf8]{inputenc} % allow utf-8 input
\usepackage[T1]{fontenc}    % use 8-bit T1 fonts
\usepackage{hyperref}       % hyperlinks
\usepackage{url}            % simple URL typesetting
\usepackage{booktabs}       % professional-quality tables
\usepackage{amsfonts}       % blackboard math symbols
\usepackage{nicefrac}       % compact symbols for 1/2, etc.
\usepackage{microtype}      % microtypography
\usepackage{xcolor}         % colors

\title{Formatting Instructions For NeurIPS 2025}

\usepackage{amsmath,amsfonts,bm}

\def\eqref#1{equation~\ref{#1}}
\def\1{\bm{1}}

\DeclareMathAlphabet{\mathsfit}{\encodingdefault}{\sfdefault}{m}{sl}
\SetMathAlphabet{\mathsfit}{bold}{\encodingdefault}{\sfdefault}{bx}{n}

\usepackage{hyperref}
\usepackage{url}
\usepackage{wrapfig}

\title{Retrieval Capabilities of Large Language Models Scale with Pretraining FLOPs}

\author{
Jacob Portes\thanks{Corresponding author: jacob.portes@databricks.com} \And
Connor Jennings \And
Erica Ji Yuen \And
Sasha Doubov \And
Michael Carbin\thanks{Joint affiliation with Databricks Mosaic Research and MIT} \And
\normalfont{Databricks Mosaic Research}
}

\usepackage{microtype}
\usepackage{graphicx}
\usepackage{subfigure}
\usepackage{booktabs} % for professional tables

\usepackage{hyperref}

\usepackage{amsmath}
\usepackage{amssymb}
\usepackage{mathtools}
\usepackage{amsthm}
\usepackage{pifont}

\usepackage[capitalize,noabbrev]{cleveref}

\theoremstyle{plain}

\theoremstyle{definition}

\theoremstyle{remark}

\usepackage[textsize=tiny]{todonotes}

\begin{document}

\maketitle

\begin{abstract}
How does retrieval performance scale with pretraining FLOPs? We benchmark retrieval performance across LLM model sizes from 125 million parameters to 7 billion parameters pretrained on datasets ranging from 1 billion tokens to more than 2 trillion tokens. We find that retrieval performance on zero-shot BEIR tasks predictably scales with LLM size, training duration, and estimated FLOPs. We also show that In-Context Learning scores are strongly correlated with retrieval scores across retrieval tasks. Finally, we highlight the implications this has for the development of LLM-based retrievers.
\end{abstract}

\section{Introduction}

% Generalized Embeddings Emerge with Scale: 

% Industry labs as well as academic research groups have invested heavily in decoder-style LLMs. A consensus has grown around scaling laws for LLMs based on perplexity and downstream in context learning (ICL) tasks, where floating point operations (FLOPs) play an important role in addition to model size and training tokens. This has led to surprisingly capable LLMs with 7-8 billion active parameters or less \citep{Touvron2023LLaMAOA,touvron2023llama,jiang2023mistral,dey2023btlm,dubey2024llama}.

%The research focus for models such as Llama 2, Llama 3, OLMo 2 \citep{olmo20242} and others has been on text generation ICL tasks such as MMLU \citep{hendrycks2020measuring} and Big-Bench \citep{srivastava2022beyond}. However, 
Recent work has shown that LLMs such as Llama 2 7B, Mistral 7B, and Mixtral 8x7B trained with causal language modeling can be naively converted into good retrieval models \citep{ma2023fine,wang2023improving}. In this study, we ask: \textbf{How well do decoder-style LLMs do on information retrieval tasks across model size, training duration, and FLOPs?}

\begin{figure*}[h]
    \centering
    \includegraphics[width=1\textwidth]{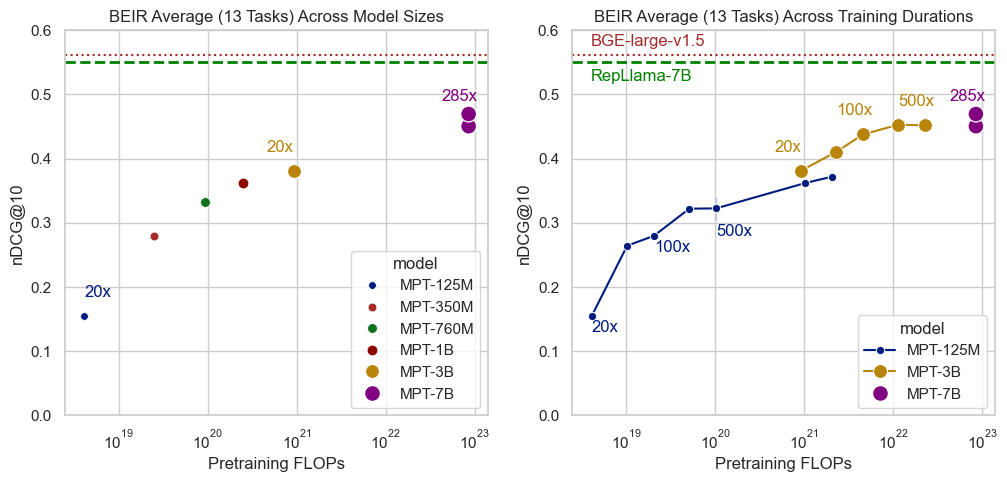}
    \caption{\textbf{Retrieval performance on BEIR improves with model size, training duration, and FLOPs.} (A) Increasing model size leads to an improvement in average BEIR nDCG@10 scores. MPT-125M, 350M, 760M, 1B, 3B models were trained for token parameter ratio of 20 on the same dataset, while MPT-7B models were trained for a token parameter ratio of 285 (i.e. a total of 2 trillion tokens). (B) Training models on more tokens leads to improvements in average BEIR score. Different MPT-125M models were trained for 3B to 1.5T tokens (20-10,000 token parameter ratios, blue line), while MPT-3B models were trained for 50B to 1.2T tokens (20-500 tokens per parameter, yellow line). All pretrained checkpoints were finetuned on 500k samples from MS MARCO with a maximum sequence length of 128 tokens. We include BGE-large-v1.5 \citep{xiao2023c} and RepLlama-7B \citep{ma2023fine} performance for comparison.}
    \label{fig:beir-average-plots}
\end{figure*}

The idea of taking pretrained checkpoints and converting them into good embedding models has been around since the early days of neural/dense retrieval.\footnote{Throughout this study, we use the terms ``neural retrieval,'' ``dense retrieval,'' and ``embedding models'' interchangeably.} For example, SentenceBERT successfully trained BERT models to embed sentence pairs \citep{reimers2019sentence}. Models such as GTR \citep{ni2021large} and Instructor \citep{su2022one} are initialized with the encoder weights from T5 models, and E5 is initialized with pretrained BERT weights \citep{wang2022text}. 

%So how do you take pretrained models and turn them into good retrieval models? 
In most of these cases, the models were further pretrained and/or finetuned with contrastive loss (a.k.a. InfoNCELoss), given the importunate fact that pretrained encoders and decoders are poor retrievers ``out of the box.'' The approach of models like E5-base and E5-large (110M and 340M parameters respectively) is to continue pretraining on millions of query-passage pairs followed by finetuning on curated datasets such as MS MARCO \citep{bajaj2016msmarco}. RepLlama on the other hand simply finetuned Llama 2 7B on 500,000 query-passage pairs from MS MARCO and found surprisingly good performance on zero-shot retrieval benchmarks such as BEIR \citep{ma2023fine,thakur2021beir}.

What is it about foundation models that allow them to perform well at information retrieval tasks? There is a recent trend where embedding models built on top of 7B parameter decoders have overtaken embedding models built on top of BERT-style models on benchmarks such as MTEB \citep{muennighoff2023mteb}. Foundation models like Llama 2 7B have been trained on 2 trillion tokens, can handle long context lengths and have strong in context learning capabilities; it is unsurprising that these new generation of LLMs do well on retrieval tasks.

Given the hard lesson that scale is (almost) all you need, \textbf{we set out to determine how retrieval performance is related to model size, training duration, and floating point operations (FLOPs)}.

In this study we explore the relationship between LLM model size, pretraining duration, and FLOPs for retrieval tasks. We start with pretrained checkpoints of MPT decoders \citep{MosaicML2023IntroducingMPT7B, sardana2023beyond} ranging in size from 125 million parameters to 7 billion parameters. These checkpoints were trained on datasets ranging from 1 billion tokens to more than 2 trillion tokens spanning token to parameter ratios of 20 through 500 (and in one case 10,000). We minimally finetune each model checkpoint for one epoch of 500,000 MS MARCO samples with InfoNCE loss using contrastive pairs with hard negatives. We then analyze zero-shot retrieval performance on the BEIR benchmark \citep{thakur2021beir}.  We find that:
\begin{itemize}
    \itemsep0em 
    \item Retrieval performance scales both with increasing model size and increasing pretraining duration for a fixed model size. This relationship is best captured by accuracy-FLOPs curves; i.e. \textbf{retrieval performance scales with pretraining FLOPs}.
    \item For most of the BEIR retrieval tasks, training a small model for more tokens has similar accuracy to a larger model trained on fewer tokens up to a ceiling. Another way of stating this is that \textbf{isoFLOPs curves for fixed model sizes significantly overlap}.
    \item In Context Learning (ICL) scores are strongly correlated with retrieval scores across BEIR tasks. Almost without exception, LLMs that have higher ICL scores also have higher retrieval scores.
\end{itemize}

The goal of this study is not to achieve state of the art retrieval performance; it is simply to investigate scaling properties as a function of tokens seen during pretraining, which has not been extensively detailed in prior work.

Our results shed light on why high quality decoders ranging in size from 1B-8B active parameters are strong candidates for embedding models. We believe this has important implications for the next generation of dense retrieval models.

%We treat the token parameter ratio as an important value in our experiments.
%\textit{How well do LLMs do at retrieval tasks across model sizes and training durations?}

\section{Related Work}

\subsection{Embedding/Dense Retrieval Models}

The idea of taking pretrained LLM checkpoints and converting them into embedding models has been around since the early days of neural/dense retrieval. Dense retrieval quickly grew in popularity after the release of BERT; for example, SentenceBERT successfully trained BERT models to embed sentence pairs \citep{reimers2019sentence}. Early approaches focused on re-ranking \citep{nogueira2019passage}, although ``full ranking'' (i.e. embedding models) shortly followed \citep{khattab2020colbert,karpukhin2020dense,izacard2021unsupervised}.

\citet{ni2021large} showed that increasing LLM model size while keeping the \textit{final embedding dimension fixed} led to improvements in zero-shot BEIR retreival performance. Specifically, in order to build their GTR models, they used the encoder half of T5-Base (110M), large (335M), XL (1.24B) and XXL (4.8B) and further pretrained them on 2 billion community question-answer pairs and then finetuned them on MS MARCO \citep{bajaj2016msmarco}. Our work builds on this direction by using decoders instead of encoders and by avoiding the ``further pretraining'' stage altogether.

%TO DO Instructor based on GTR (in turn based on T5), inspired by instruction finetuning, focused heavily on finetuning with instructions. Discuss instruction finetuning in more detail

Many studies have shown that BERT-base and BERT-Large size models can achieve state-of-the-art performance on retrieval benchmarks such as BEIR when trained with contrastive pairs \citep{wang2022text, xiao2023c}. \citet{wang2022text} trained E5 on millions of contrastive pairs with soft negatives but did not release their pretraining data, while BGE \citep{xiao2023c} followed a similar recipe and gave more hints as to their pretraining data sets.

\citet{ma2023fine} finetuned Llama 7B weights on MS MARCO and found very good performance on BEIR. Our work builds on RepLlama by exploring the scaling properties of LLM-based retrieval models.
Some of the same authors as RepLlama and E5 achieved state of the art performance on BEIR by finetuning Mistral-7B-Instructor \citep{jiang2023mistral} on 13 datasets of contrastive pairs with hard negatives as well as synthetic contrastive pairs generated by GPT-3.5/4  \cite{wang2023improving}.

There has been a recent explosion of LLM-based embedding models such as SFR-Embedding-Mistral \citep{SFRAIResearch2024}, NV-Embed \citep{lee2024nv}, bge-en-icl \citep{li2024making}, gte-Qwen2-1.5B-instruct \citep{li2023towards}, GRIT LM \citep{muennighoff2024generative} and others. The Qwen 3 Embedding models (0.6B, 4B 8B) models were initialized from the Qwen 3 pretrained models \citep{zhang2025qwen3}. Similarly, the Gemini Embedding models \citep{lee2025gemini} were initialized from the Gemini models \citep{team2023gemini}. There has also been a recent uptick in BERT-style embedding models including Nomic Embed \citep{nussbaum2024nomic}, Arctic-Embed \citep{merrick2024arctic}, and ModernBERT \citep{warner2024smarter}. 

\subsection{Evaluation of Embedding Models}

Neural network based embedding and retrieval were considered slightly different subfields. Over time these have converged, as exemplified by the MTEB benchmark \citep{muennighoff2023scaling}, which incorporates BEIR \citep{thakur2021beir} as well as many other benchmarks for tasks such as semantic similarity. We give more details on the BEIR and MTEB benchmarks below. More recent retrieval benchmarks include AIR Bench \citep{chen2024air}, LongEmbed \citep{zhu2024longembed} and BRIGHT \citep{su2024bright}; we save the analysis of LLM scaling properties on these benchmarks for future work.

\subsection{Scaling Laws for Large Language Models}

The study by \citet{hoffmann2022training}, informally known as the ``Chinchilla paper,'' determined scaling laws for optimally allocating train compute for LLMs. They arrived at a heuristic that models should be trained on a total number of tokens that is roughly $20\times$ the number of model parameters (i.e. the token-per-parameter ratio). They use three approaches to predict optimal model size and number of tokens for pretraining a LLM with a fixed compute budget; first they fix model size and vary dataset size, and then they also establish isoFLOP profiles by varying model size for fixed number of FLOPs. Finally, they fit a parametric loss function. All approaches arrive at the same rough conclusion that LLMs such as GPT-3 \citep{brown2020language} and Gopher \citep{rae2021scaling} were significantly under trained. This built on previous work exploring scaling laws for LLMs \citep{hernandez2021scaling, kaplan2020scaling, tay2021scale}. 

The popularity - and capabilities - of models such as Llama 2 7B, Mistral 7B \citep{jiang2023mistral} and Llama 3 8B have upended the chinchilla scaling laws. These models were trained on far more tokens than ``Chinchilla optimal;'' for example, Llama 2 7B was pretrained for 2 trillion tokens, which is a token parameter ratio of 285. Recent studies have argued that inference serving costs (which increase with model size) as well as ease of use should be taken into consideration when training LLMs \citep{devries2023chinchilla_analysis, sardana2023beyond, epoch2023tradingoffcomputeintrainingandinference}.

All these studies focus on cross entropy loss and not on downstream ICL performance (or retrieval performance for that matter). While a study by \citet{tay2021scale} found that T5 model size did not follow easily identifiable scaling laws on downstream SuperGLUE performance, there is little work showing how current decoder downstream performance scales with FLOPs. There are even fewer studies that focus on the question of how LLM retrieval performance scales with FLOPs.

\citet{muennighoff2022sgpt} investigated the scaling properties of GPT for semantic similarity from 100M - 5.8B parameters.
Motivated by the promise of LLM-based retrievers, \cite{jiang2023scaling} propose an ICL-based method to improve sentence embedding performance and use it to finetune OPT models from size 125M to 66B and Llama models across 7B, 13B, 30B and 65B. In this work, however, they exclusively focus on the semantic textual similarity benchmarks. Finally, \citet{fang2024scaling} derived phenomenological scaling laws for dense embeddings based on BERT models up to 80M parameters. 
\cite{zeng2025scaling} investigate the retrieval across the Llama 3 1B, 3B and 8B models.
Our study complements these results and focuses on the scaling properties of decoders from 125M parameters up to 7B parameters with varying token per parameter ratios.

\section{Results}

We used checkpoints of pretrained MPT decoders of sizes varying from 125 million parameters to 7 billion parameters. These checkpoints were trained for various durations ranging from 1 billion tokens to 2 trillion tokens with corresponding token-per-parameter ratios of 20 through 500 (and in one case up to 10,000). Note that each checkpoint was trained to completion, with a full learning rate schedule consisting of a warmup followed by a cosine decay. See \citet{MosaicML2023IntroducingMPT7B, MosaicML2023IntroducingMPT30B} for further pretraining details.
We then minimally finetuned these models on the now-classic MS MARCO dataset for 1 epoch (500,000 samples) of query-passage pairs with hard negatives and evaluated them on the BEIR retrieval benchmark.

%\subsection{MPT Pretrained Checkpoints}

%\textcolor{red}{Rough model dimensions. MPT architecture with ALiBi, sequence length 4096, trained using next token prediction (ie causal language modeling). Trained using Composer and Streaming python libraries. GPT4 tokenizer tiktoken. See App.~\ref{app:mpt-models} for more architecture details. warmup with DecoupledAdamW, LionW. Batch size, microbatch size. We then finetune these models with a contrastive InfoNCE loss.}

\subsection{Finetuning on MS MARCO with the InfoNCELoss}

Using a pretrained LLM, we pool and average the respective tokens of the final hidden representations of each document and query.\footnote{Other approaches here include using an end of sentence token to represent the content of the preceding tokens.} To finetune the model, we leverage the class InfoNCE loss as follows over the negatives:
\begin{equation}
    \min L_{\text {cont }}=-\frac{1}{n} \sum_i \log \frac{\mathrm{e}^{s_{\boldsymbol{\theta}}\left(q_i, p_i\right)}}{\mathrm{e}^{s_{\boldsymbol{\theta}}\left(q_i, p_i\right)}+\sum_j \mathrm{e}^{s_{\boldsymbol{\theta}}\left(q_i, p_{i j}^{-}\right)}}
\end{equation} 
 
where $q_i$ is a query, $p_i$ is a ``positive'' passage that is paired with the query, and $p_{ij}^{-}$ is a ``negative'' passage that is somewhat relevant to the query but not the correct passage.\footnote{In this study we use ``hard negatives'' as opposed to ``in-batch'' negatives.} $s_{\theta}$ is the cosine similarity function, and $n$ is the total number of samples in a batch. This formula can be expressed in terms of cross entropy, which makes for easy implementation in PyTorch. Our implementation was motivated by \citet{wang2022text}.\footnote{\url{https://github.com/microsoft/unilm}}

%Fully Shared Data-Parallel (FSDP) is a widely adopted technique for multi-machine training. 
%Our InfoNCE implementation supports distributed training, and uses batches from across all machines as in-batch negatives for a given query.

The positive document for a given query is derived from the MS MARCO Document Retrieval dataset \citep{bajaj2016msmarco}.  We use 15 curated hard negative documents per query, mined using BM25 \citep{robertson1995okapi}.\footnote{We note that as of 2025, there are more effective ways to mine for hard negatives than BM25.}

%\textcolor{blue}{TO DO: Discuss how big MS MARCO dataset is and how it is formatted. Also what version we used. Maybe an example}

%Batch size, microbatch size, warmup, learning rate, etc.

While all MPT models were pretrained with a maximum sequence length of 4096 tokens, we finetuned and evaluated all models with a maximum sequence length of 128 tokens.\footnote{This was due to compute constraints. We note that this is far below current industry standards.} We also used the same hyperparameters such as warmup and learning rate for all model sizes without doing hyperparameter sweeps. This strongly handicaps the models; we therefore interpret all our BEIR scores as \textit{lower bounds} on retrieval performance.

\begin{figure*}[p]
    \centering
    \includegraphics[width=1\textwidth]{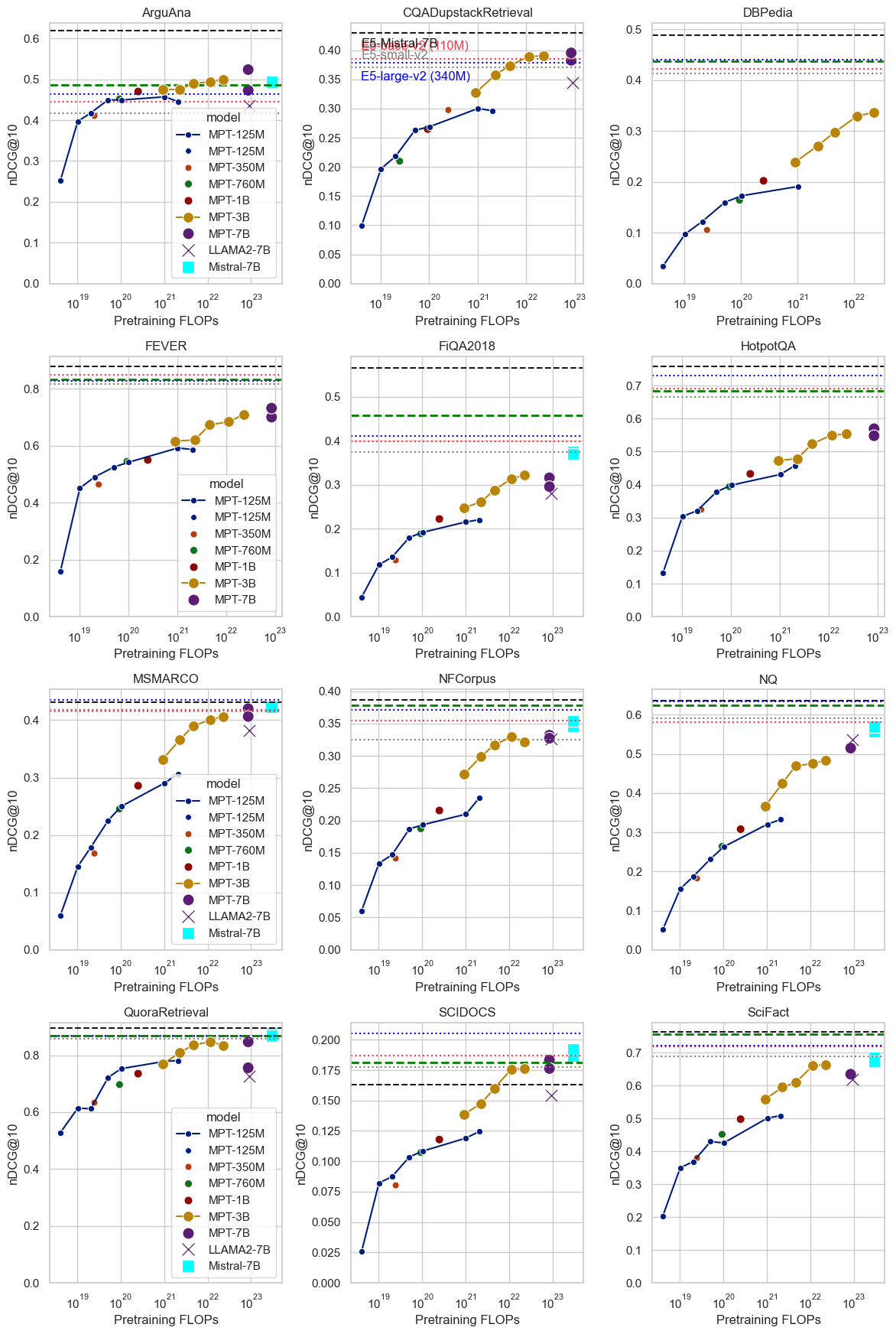}
    \caption{\textbf{LLM performance on individual BEIR tasks scales with model size, pretraining duration, and FLOPs}. E5-small-v2 (grey), E5-base-v2(red), E5-large-v2 (blue), E5-Mistral-7B (black), and RepLlama-7B (green) performance is included for reference.}
    \label{fig:beir-individual-plots}
\end{figure*}

\subsection{Estimating FLOPs}
Floating point operations (FLOPs) is a hardware-agnostic metric that conveys how much ``compute'' was used to train a particular model.
To first order, FLOPs for dense transformer models can be estimated as $ 6 \times N\times tokens $ where $N$ is the total number of model parameters \citep{kaplan2020scaling,chowdhery2023palm,transformer-math-eleutherai}. Thus FLOPs increase with both training duration (i.e. data) and model size. We use this approximation for all FLOPs estimates.

\subsection{BEIR Retrieval Benchmark}

We describe the MTEB \cite{muennighoff2023mteb} version of BEIR \citep{thakur2021beir}, which we used for all of our evaluations.

%\subsubsection{BEIR Retrieval Tasks}

% JP NOTE: see this notion page for stats on the different tasks https://www.notion.so/Understanding-the-BEIR-Tasks-October-2023-February-2024-ecca99c8794e449f917196f1828168fa?pvs=4

Each benchmark within BEIR is divided into queries and documents (a.k.a ``passages''), and the task is to find most relevant documents for a given query. Exact search is done using cosine similarity (as opposed to approximate search).  SCIDOCS is one example dataset that contains 1,000 queries, and 25,657 documents from scientific publications in the test set \citep{cohan2020specter}. The various BEIR tasks are detailed in the \Cref{appendix:beir}. 

BEIR was originally designed to be a zero-shot evaluation benchmark, which means that many of the early retrieval models were careful \textit{not} to train in-domain. However, all of the ``top'' embedding models on the MTEB benchmark not only train on the training sets associated with each BEIR benchmark, but also arguably train on the corpus test sets of each BEIR task \citep{wang2022text,wang2023improving,xiao2023c}.
Since many of the benchmarks include common datasets derived from Wikipedia, Stack Exchange, many argue that it is unrealistic to treat BEIR as a zero-shot evaluation benchmark. Instead of focusing on benchmark hacking, in this study we choose to benchmark the scaling properties of pretrained decoders by minimally finetuning on a single classic embedding dataset (MS MARCO). We hope that future work will use more curated, higher-quality finetuning datasets to show similar scaling properties.

%\Table{}
%Dataset | # queries | # passages | dev or test? | task description

%\begin{figure}

\subsection{Retrieval Performance Scales with Model Size, Training Duration, and FLOPs}

We first evaluated MPT 125M, 350M, 760M, 1B and 3B models all pretrained to a token per parameter ratio of 20, ranging from roughly 3 billion tokens to 50 billion tokens (see Appendix Table \ref{tab:training_duration}).
Figure \ref{fig:beir-average-plots}A shows the smooth increase of average BEIR score with model size as a function of pretraining FLOPs.

We then used MPT-125M model checkpoints for various training durations ranging from 3 billion tokens, i.e. 20 tokens per parameter, through 1.5 trillion tokens, i.e. an extreme of 10,000 tokens per parameter (Figure  \ref{fig:beir-average-plots}B blue line). Somewhat surprisingly, \textit{we find that retrieval performance steadily improves with increased training duration and does not plateau}.\footnote{This result is similar to \cite{sardana2023beyond}, which used many of the same checkpoints.} We also used pretrained MPT-3B model checkpoints for various training durations ranging from 50B i.e. 20 tokens per parameter, to 1.2 trillion tokens, i.e. 500 tokens per parameter. Here too we see a steady increase in retrieval performance as a function of FLOPs. Finally, we also finetuned two slightly different MPT 7B checkpoints pretrained on 2T tokens with a token per parameter ratio of 285 (\Cref{fig:beir-average-plots} purple dots). As expected, these have slightly higher average nDCG@10 than the MPT-3B checkpoint trained to 500 tokens per parameter. 

Figure \ref{fig:beir-individual-plots} shows scores for individual BEIR tasks as a function of pretraining FLOPs. We show Llama 2 7B and Mistral 7B performance on some tasks for comparison (finetuned in the same manner as all of the other checkpoints).\footnote{Note here that the Mistral 7B pretraining FLOPs are unknown; we simply assume that Mistral was trained on slightly more data than Llama 2 7B.} As expected, the scaling trends for individual BEIR tasks are the same as in Figure \ref{fig:beir-average-plots}, with the notable exceptions of Arguana and Touche 2020 (not shown). For Arguana, performance essentially plateaus despite increasing FLOPs. This is likely due to the unusual nature of the task, which requires the model to find a \textit{counterargument} for a given argument text (i.e. the query). The only models that excel at this task are models that explicitly include \textit{instructions to find the counterargument} in the query, like E5-Mistral-7B \citep{wang2023improving}.\footnote{See Appendix Table 14, where the evaluation instructions are ``Given a claim, find documents that refute the claim''} We speculate on the role of explicit instructions in the Discussion.

\begin{wrapfigure}{r}{0.5\textwidth}
    \centering
    \includegraphics[width=0.5\textwidth]{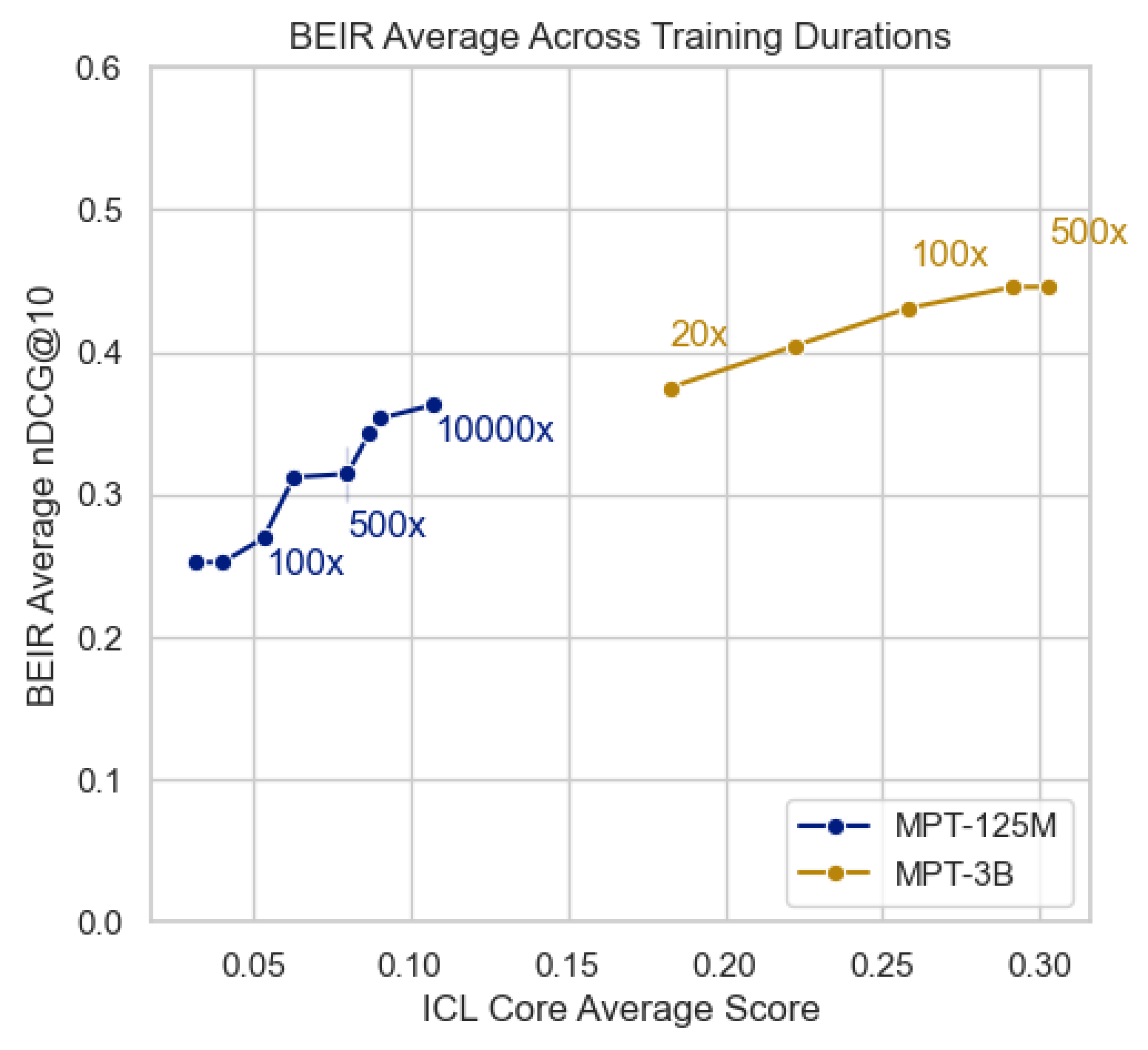}
    \caption{\textbf{Retrieval performance is highly correlated with In Context Learning (ICL) performance}. ICL Core Average Score uses the open-source MosaicML Evaluation Gauntlet (39 tasks).}
    \label{fig:average-icl}
\end{wrapfigure}

\subsection{Small models trained on more data can match performance of larger models trained on less data}

One surprising observation from \Cref{fig:beir-average-plots} and \Cref{fig:beir-individual-plots} is that small models pretrained on more data (i.e. for more FLOPs) can match the performance of larger models trained on less data. For example, MPT-125M trained for 1.5T tokens at roughly a 10,000 token parameter ratio has same average BEIR performance as a MPT-1B model trained for 25.2B tokens with a token parameter ratio of 20.

\subsection{BEIR scores are strongly correlated with ICL Scores }

We then plot the BEIR nDCG@10 scores as a function of the ICL scores for each pretrained checkpoint using the open-source MosaicML Evaluation Gauntlet (39 tasks) \citep{Dohmann2023,Barton2024}. Figure \ref{fig:average-icl} demonstrates that ICL scores of pretrained checkpoints are strongly correlated with BEIR scores after contrastive finetuning on MS MARCO. This provides strong evidence that BEIR is a good measure of LLM retrieval capability. See Appendix \ref{fig:icl-beir-plot-individual} for the nDCG vs. ICL performance for individual BEIR tasks. 

There is a noticeable gap in average ICL score between MPT-125M with token parameter ratio 10,000 (blue line) and MPT-3B with token per parameter ratio 20 (yellow) along the x-axis for all tasks. This simply indicates that MPT-3B models are always better than MPT-125M models for the ICL tasks. We note that this gap appears for ICL tasks but not for retrieval tasks, and interpret this to mean that retrieval tasks are easier than ICL tasks.\footnote{One can also argue based on this that one can get more mileage on retrieval tasks from smaller models trained for longer.}

% \section{Results}

% \subsection{A General Recipe for Turning an LLM into an Embedding Model}

% The recipe is: contrastive finetuning on 500k generic Query-Passage pairs
% We show that it works for different model sizes (MPT-110M, MPT-340M, MPT-1B, MPT-3B, MPT-7B) and for different model classes (Llama-7b, Llama-13b).

% \subsection{In-Domain LLMs can Also be Used for In-Domain Embeddings}

% Here we show that his approach works for LLMs that have been further pretrained for specific domains.

% \subsection{Data Mix Experiments}

% In this section we give details on our experiments invovling different mixtures of contrastive datasets.
% \begin{itemize}
%     \item contrastive pairs with hard negatives vs. no hard negatives
%     \item different combinations of MS MARCO, NQ, etc.
% \end{itemize}

% \subsection{Using the same LLM and LLM-Embed variant improves RAG performance}
%  In this section we show that using MPT-Embed with MPT in a RAG system is better than using a generic embedding model with MPT in a RAG system.

\section{Discussion}

Our work shows that retrieval performance scales with LLM pretraining FLOPs, and we hope that it provides strong motivation for finetuning pretrained LLMs for retrieval. 

Why might one want to build an embedding model using a pretrained LLM? There has been a cambrian explosion of open-source LLMs in the range of 1B, 3B, 7B, 8B, and 13B parameters (as well as Mixture of Experts models with a similar number of active parameters).\footnote{Llama 2 7B and 13B, Llama 3.2 1B and 3B, Gemma 2 3B, Llama 3.3 8B, \href{https://huggingface.co/mistralai/Mistral-7B-v0.3}{Mistral 7B}, \href{https://huggingface.co/mistralai/Mixtral-8x7B-Instruct-v0.1}{Mistral 8x7B}, OLMo 2 7B and 13B, \href{https://huggingface.co/allenai/OLMoE-1B-7B-0924}{OLMoE-1B-7B}, \href{https://huggingface.co/deepseek-ai/deepseek-moe-16b-base}{DeepSeekMoE-3B-16B}.} Almost all of these models have been pretrained on \textit{trillions} of tokens, can handle long context lengths upwards of 8k tokens, and have been extensively finetuned to handle nuanced language using both supervised finetuning and reinforcement learning.

The previous generation of embedding models such as E5, GTE and BGE were built on top of pretrained BERT models. BERT is 6 years old and counting, and there is much less development occurring for small encoder models.\footnote{With the minor exceptions of MosaicBERT \citep{portes2023mosaicbert}, CrammingBERT \citep{geiping2023cramming}, NomicBERT \citep{nussbaum2024nomic}, and ModernBERT \citep{warner2024smarter}.} While BERT-style models are mostly smaller than 1B parameters and therefore easier to deploy, 7B models are already dominating retrieval benchmarks such as MTEB. We suspect that this trend away from BERT models will continue.

Does retrieval performance continue to increase beyond 7B models? While there are mixed reports on the success of retrieval models larger than 7B parameters (e.g. RepLlama 13B, Mixtral 8x22B), our results hint at the tantalizing possibility that retrieval performance might continue to scale. With the recent state of the art open weights models such as DeepSeek, Qwen 3, and Llama 4, we expect research on decoder-based retrieval to continue in this direction. 

Why don't any of our checkpoints achieve SOTA on BEIR or MTEB? In this study, we chose to focus on the scaling properties of pretrained models using a baseline finetuning approach. All of the datapoints represented by circles were finetuned on the same MS MARCO dataset with the same hyperparameters and a maximum sequence length of 128 tokens. This is a \textit{lower bound} of performance, as most modern embedding models finetune on a maximum sequence length of longer than $512$ tokens and use much more extensive finetuning datasets.

One example of a more extensive finetuning dataset is E5-Mistral-7B \citep{wang2023improving}. The E5-Mistral-7B team finetuned Mistral-7B on a custom, closed source hard-negatives dataset that consists of 13 semi-open-source datasets (they mined their own hard negatives) including MS MARCO as well as synthetically generated GPT-3.5/4 data \citep{wang2023improving}. We only trained our checkpoints on MS MARCO.

In this study, we don't derive phenomenological scaling laws; rather we report a strong trend. We save formal retrieval scaling laws for future work.

\section*{Acknowledgements}

We'd like to thank Andrew Drozdov, Vitaliy Chiley, Nikhil Sardana, and Michael Bendersky for feedback on this work.

\section*{Contributions}

JP led this project, ran experiments, and wrote this manuscript.  CJ and EJY ran early finetuning experiments. SD ran some of the pretraining experiments. MC advised this work.

The entirety of this paper was written without AI =)

%Future directions:
%need for benchmarks with longer context such as LongEmbed \citep{zhu2024longembed}
%ColBERT hasn't been tested at scale
%Retrieval Augmented Generation

%ColBERT is an alternative approach to single vector embeddings that exhibits good out-of-domain generalization behavior \cite{khattab2020colbert,santhanam2021colbertv2}. Instead of representing the query and passage as single vectors, ColBERT preserves the final vectors for each token in the query and passage and compares each token with every other token using a maxSim operation. This ``late-stage interaction'' does however lead to computational overhead. Additionally, ColBERT models are usually smaller than 340 million parameters.  Seeing how the ColBERT approach works for larger model sizes is an interesting direction for future research.

%\textbf{Do not} include acknowledgements in the initial version of the paper submitted for blind review.

% \section*{Impact Statement}

% This paper presents work whose goal is to advance the field of Machine Learning. There are many potential societal consequences of our work, none which we feel must be specifically highlighted here.


\clearpage
\bibliography{neurips_2025}
\bibliographystyle{iclr2025_conference}
%\bibliographystyle{iclr_fmwild}

%%%%%%%%%%%%%%%%%%%%%%%%%%%%%%%%%%%%%%%%%%%%%%%%%%%%%%%%%%%%%%%%%%%%%%%%%%%%%%%
%%%%%%%%%%%%%%%%%%%%%%%%%%%%%%%%%%%%%%%%%%%%%%%%%%%%%%%%%%%%%%%%%%%%%%%%%%%%%%%
% APPENDIX
%%%%%%%%%%%%%%%%%%%%%%%%%%%%%%%%%%%%%%%%%%%%%%%%%%%%%%%%%%%%%%%%%%%%%%%%%%%%%%%
%%%%%%%%%%%%%%%%%%%%%%%%%%%%%%%%%%%%%%%%%%%%%%%%%%%%%%%%%%%%%%%%%%%%%%%%%%%%%%%
\newpage
\appendix
\onecolumn

\begin{table}
    \centering
    \begin{tabular}{c c c c c c } \toprule 
         Model Name&  Total Parameters&  Hidden Dimension &  Attention Heads & Layers & Expansion Ratio \\ \toprule 
         MPT-125M&  151M& 768 & 12 & 12 & 4   \\ 
         MPT-350M&  367M& 1024 & 16 & 24 & 4   \\ 
         MPT-760M&  749M& 1536 & 12 & 24  & 4   \\  
         MPT-1B& 1.26B& 2048 & 16  & 24 & 4\\ 
         MPT-3B& & 2560 & 20 & 32 & 4\\ 
         MPT-7B& & 4096 & 32 & 32 & 4\\ \bottomrule
         %Llama-7B&  &  &  &  & \\ 
    \end{tabular}
    \caption{MPT Model Architecture Details}
    \label{tab:model_architecture}
\end{table}

\begin{table}
    \centering
    \begin{tabular}{l c c c c c c c} \toprule 
         TPR &  MPT-125M &  MPT-350M &  MPT-760M & MPT-1B & MPT-3B & MPT-7B &  Llama-7B\\ \toprule 
         20    & 3.02B & 7.34B  & 14.98B   & 25.2B    & 49.2B    & -     & -     \\
         50    & 7.55B & 18.35B & -        & 63B      & 123B     & -     & -     \\
         100   & 15.1B & 36.7B  & 74.9B    & 126B     & 246B     & 700B  & -     \\
         250   & 37.8B & 73.4B  & 187.25B  & -        & 615B     & -     & -     \\
         285   & -     & -      & -        & -        & -        & 2000B & 2000B \\
         500   & 75.5B & 183.5  & 374.5B   & -        & 1230B    & -     & -     \\
         1000  & 151B  & -      & -        & -        & -        & -     & -     \\
         5000  & 755B  & -      & -        & -        & -        & -     & -     \\
         10000 & 1510B & -      & -        & -        & -        & -     & -     \\
         \bottomrule
    \end{tabular}
    \caption{Training Duration (tokens). Note that one MPT-125M checkpoint was trained for 1.5 trillion tokens.}
    \label{tab:training_duration}
\end{table}

\section{Retrieval metrics}
While there are many metrics used for retrieval, normalized discounted cumulative gain (nDCG) is a particularly common one. This metric indicates whether the retrieved documents are (1) relevant and whether (2) they are sorted in order of relevance to the query. Specifically, it is defined as: $$ \mathrm{DCG}_{\mathrm{p}}=\sum_{i=1}^p \frac{r e l_i}{\log _2(i+1)}=r e l_1+\sum_{i=2}^p \frac{r e l_i}{\log _2(i+1)} $$
$$\mathrm{IDCG}_{\mathrm{p}}=\sum_{i=1}^{\left|R E L_p\right|} \frac{r e l_i}{\log _2(i+1)}$$
where $rel_i$ is graded relevance of the result at position $i$. Finally, the normalized discounted cumulative gain is defined as:

\begin{equation}
    \mathrm{nDCG}_{\mathrm{p}}=\frac{D C G_p}{I D C G_p}
\end{equation}

\section{BEIR Retrieval Tasks}
\label{appendix:beir}

\textbf{ArguAna} \cite{wachsmuth2018retrieval}: 1,406 queries and 8,674 documents in the test set. Data consists of single sentence ``arguments'' and single sentence ``counterarguments'' originally curated from the obscure online debate portal idebate.org. The task is to find documents that \textit{refute} the claim in the query, which makes the task particularly difficult without instructions or instruction tuning.

\textbf{ClimateFEVER} \cite{diggelmann2020climate}: similar in spirit to FEVER, ClimateFEVER is a dataset for verification of climate change-related claims.

\textbf{CQADupstackRetrieval} \cite{hoogeveen2015cqadupstack}. The task is designed such that, given a single sentence title (the query), the model has to retrieve a duplicate document (title + body). 

\textbf{DBPedia} \cite{hasibi2017dbpedia} is derived from Wikipedia pages.

\textbf{FEVER} \cite{thorne2018fever} The original Fact Extraction and VERification dataset was collected semiautomatically as part of an automatic fact checking task. The task is to retrieve Wikipedia abstracts that support or refute a given claim.

\textbf{FiQA2018} \cite{maia2018fiqa}: 648 queries and 57,638 documents in the test set. passages were scraped from StackExchange posts under the Investment topic from 2009-2017. 

\textbf{HotpotQA} \cite{yang2018hotpotqa}: 5,447 queries. Multi-hop like questions derived from Wikipedia that require multiple separate paragraphs to answer. 

\textbf{MSMARCO} \cite{bajaj2016msmarco}: Since all models in this study were finetuned on MS MARCO, this is not considered ``zero-shot.''

\textbf{NFCorpus} \cite{boteva2016full}: 323 queries and 3633 documents in the test set. Queries taken from Nutrition Facts website, annotated medical documents from PubMed are the documents.

\textbf{NQ} \cite{kwiatkowski2019natural}: natural questions is Google searches with answer spans from Wikipedia articles. 3,452 queries, search over 2,681,468 passages. While the original NQ dataset includes full articles, the MTEB BEIR version only uses wikipedia abstracts \textit{and not full articles}.

\textbf{QuoraRetrieval}: 10,000 queries, and 522,931 queries as ``documents'' in the test set. The task is to find matching/duplicate queries.

\textbf{SCIDOCS} \cite{cohan2020specter} consists of scientific documents.

\textbf{SciFact} \cite{wadden2020fact} contains roughly 1,400 expert-written scientific claims. These are paired with paper abstracts.

\textbf{Touche2020} \cite{bondarenko2020overview}: 49 queries, 382,545 documents in the test set. conversational arguments. Use conclusion as title and premise as arguments. Note that only 49 queries is potentially quite noisy.

\textbf{TREC-COVID} \cite{voorhees2021trec}: 50 queries and 171,332 documents in the test set. This consists of data from the CORD-19 dataset of published scientific articles dealing with the COVID-19 pandemic.

% \section{MPT Models Supplement.}
% \label{app:mpt-models}

% MPT Model specifications
% Explain what all the the values in the table mean

% Vocab size for MPT models: 100352, vocab size for Llama models: 32k
% Context length

% % BL: In updated versions of this paper, reminder to add that most of these models use grouped query attention but the 7B models do not.
% % BL: Can remove expansion ratio from this table beacuse it's the same across models.

% Table showing number of tokens of training for different TPR
% Tabel of datasets

% \section{In-Context Learning Evaluation Gauntlet}

% TO DO Describe our eval gauntlet v2 and link to code

% \section{Correlation Between In Context Learning and Retrieval Tasks}

% Individual plots below. Could also plot ICL performance

% There is a noticeable gap in average ICL score between MPT-125M with token parameter ratio 10,000 (blue line) and MPT-3B with token per parameter ratio 20 (yellow) along the x-axis for all tasks. This simply indicates that MPT-3B models are always better than MPT-125M models for the ICL tasks. There is also a gap for retrieval accuracy along the y-axis (except for QuoraRetrieval).

\begin{figure*}[p]
    \centering
    \includegraphics[width=0.8\textwidth]{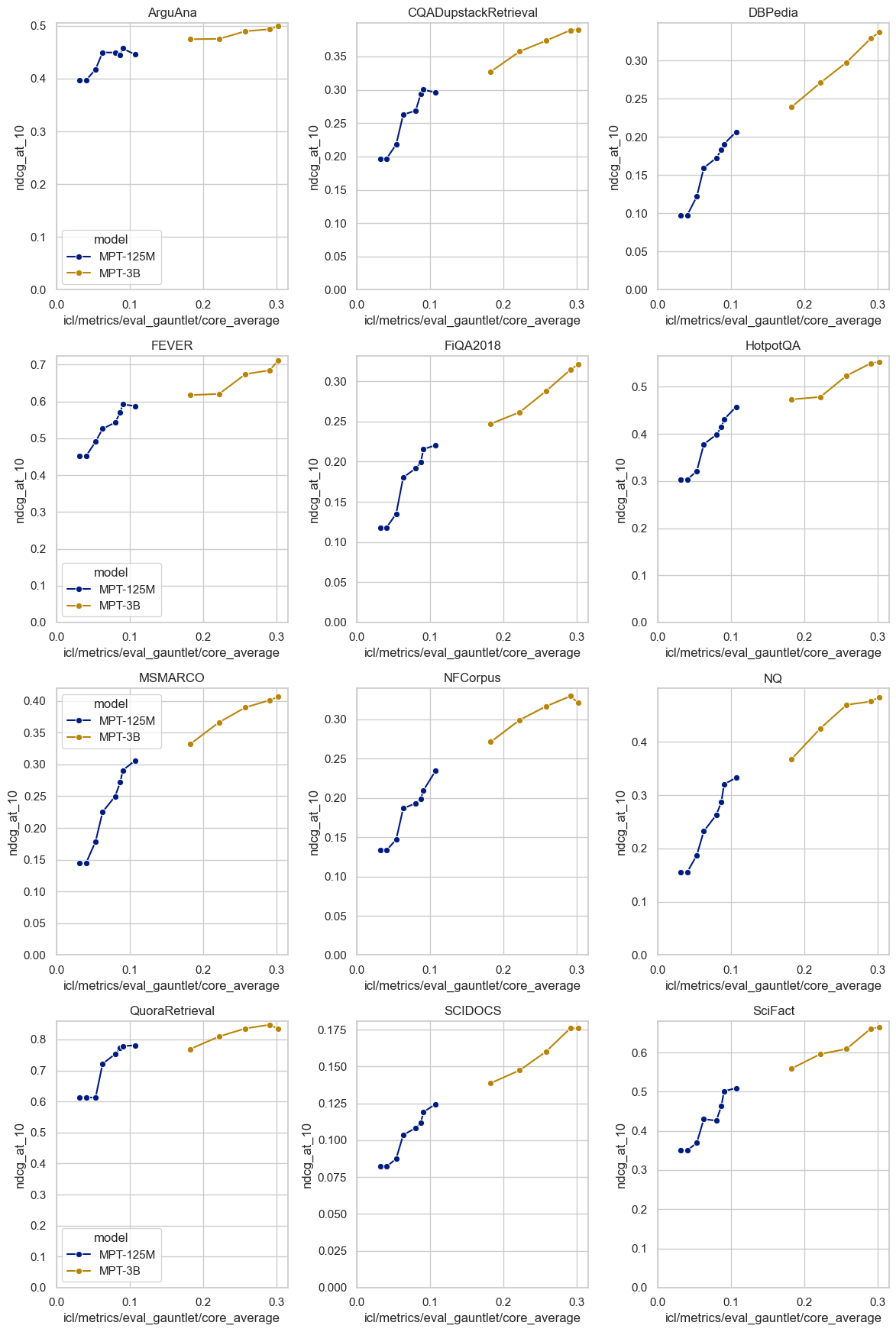}
    \caption{\textbf{Retrieval performance on individual BEIR tasks is highly correlated with ICL performance.} ICL Core Average Score uses the open-source MosaicML Evaluation Gauntlet (39 tasks). Same data as \Cref{fig:average-icl}.}
    \label{fig:icl-beir-plot-individual}
\end{figure*}

% \center
% \begin{tabular}{lr}
% \toprule
% \textbf{Parameter}  & \textbf{MPT-7B} \\ \midrule
% \texttt{d_model}             & $4096$           \\
% \texttt{n\_heads}       & $32$             \\
% \texttt{n\_dim}       & $128$            \\
% \bottomrule
% \label{tab:param}
% \end{tabular}

%%%%%%%%%%%%%%%%%%%%%%%%%%%%%%%%%%%%%%%%%%%%%%%%%%%%%%%%%%%%%%%%%%%%%%%%%%%%%%%
%%%%%%%%%%%%%%%%%%%%%%%%%%%%%%%%%%%%%%%%%%%%%%%%%%%%%%%%%%%%%%%%%%%%%%%%%%%%%%%

% This document was modified from the file originally made available by
% Pat Langley and Andrea Danyluk for ICML-2K. This version was created
% by Iain Murray in 2018, and modified by Alexandre Bouchard in
% 2019 and 2021 and by Csaba Szepesvari, Gang Niu and Sivan Sabato in 2022.
% Modified again in 2023 and 2024 by Sivan Sabato and Jonathan Scarlett.
% Previous contributors include Dan Roy, Lise Getoor and Tobias
% Scheffer, which was slightly modified from the 2010 version by
% Thorsten Joachims & Johannes Fuernkranz, slightly modified from the
% 2009 version by Kiri Wagstaff and Sam Roweis's 2008 version, which is
% slightly modified from Prasad Tadepalli's 2007 version which is a
% lightly changed version of the previous year's version by Andrew
% Moore, which was in turn edited from those of Kristian Kersting and
% Codrina Lauth. Alex Smola contributed to the algorithmic style files.

\end{document}